\documentclass[10pt,twocolumn]{article}

\usepackage[letterpaper,top=0.66in,bottom=0.72in,left=0.66in,right=0.66in,
            columnsep=0.25in,headsep=0.18in]{geometry}
\usepackage{newtxtext,newtxmath}

\usepackage{amsmath,amssymb}
\usepackage{graphicx}
\usepackage{booktabs,tabularx}
\usepackage{microtype}
\usepackage{authblk}
\usepackage{caption}
\usepackage{natbib}
\usepackage[hidelinks]{hyperref}

\hypersetup{
  pdftitle={Forged Peer Judgments Mislead Multimodal LLM Judge Panels: Source-Blind Anchoring and Panel-Consensus Verification},
  pdfauthor={Yang Shu}
}
\title{Forged Peer Judgments Mislead Multimodal LLM Judge Panels: Source-Blind Anchoring and Panel-Consensus Verification}
\author[1]{Yang Shu\thanks{Corresponding author: \href{mailto:shuyang@zju.edu.cn}{shuyang@zju.edu.cn}.}}
\affil[1]{Zhejiang University, Hangzhou, China}
\date{\small Preprint --- August 2026}

\begin{document}

\maketitle

\begin{abstract}
Multimodal LLM judge panels can cross-reference peers, but a quoted peer judgment may itself be untrusted. We expose \emph{source-blind anchoring} as a text-level attack surface in vision-language model (VLM) panels. Quoting independent visual judgments creates large anchoring gaps (19--26 percentage points) under both self and peer framing. A matched-content, label-only control changes the broken rate by only $-0.17$pp (95\% CI $[-0.68,0.35]$), showing that the self/peer label itself does not explain the effect. Under our tested construction, deliberately generated, concise wrong quotes overturn originally-correct verdicts 1.5--2.7$\times$ more often than naturally occurring wrong peer statements, with bootstrap 95\% CIs excluding parity across two datasets and seven VLM judges. Because the two statement populations differ in selection and form, this ratio measures differential damage under the tested attack rather than a provenance-only causal effect. We then introduce panel-consensus verification, which cross-checks a quote against independently collected blind votes. It blocks 84.9\% of fabricated attacks, cuts their net harm by 97.5\%, and preserves the positive but statistically inconclusive point estimate for genuine peer information under leave-one-out re-verification. These results identify a low-cost attack surface and a concrete defense for safer multimodal collaborative evaluation.
\end{abstract}

\section{Introduction}

A quoted peer judgment is useful only if the panel can trust what was quoted.

Large language model (LLM) judges are widely used to evaluate AI-system outputs at scale. As evaluation expands from text to image-grounded settings, a natural collaborative design is a panel of vision-language model (VLM) judges that can consult one another's independent assessments of the same image. The intuition is appealing: several independent viewpoints ought to be more robust than any single judgment.

This ``cross-referencing'' design, however, leaves a premise largely unexamined: how do judges respond when the quoted ``peer judgment'' is untrusted? We first establish that VLM judge panels exhibit correlated errors, consistent with existing findings on text-only judge panels -- most notably \citet{kohli2026nine}'s report that supplying a panel with the ground-truth answer closes at most 11\% of the resulting reliability gap. Using an operational taxonomy, we label 61.5\% of errors Type A (language-prior-consistent), 22.5\% Type C (visual-anchoring-failure-consistent), and 3.0\% Type B (visual-hallucination-consistent). The Type-B cell is small ($n=49$) and its pseudo-ground-truth reproducibility is uncertain; these labels are diagnostic correlates rather than proof of unique causal origin.

We then test anchoring with a blind-first protocol: before scoring, a judge independently answers a probe targeting where the candidate responses disagree, and that answer is later quoted before the final verdict. C5v2 quotes the judge's own answer under self framing; C6 quotes a different panel member's answer under peer framing. Because the quoted contents also differ, this comparison is not label-only. Nevertheless, both conditions show large broken-vs-kept correctness gaps (26.4 and 18.9pp; both bootstrap CIs exclude zero), establishing that peer statements can anchor nearly as strongly as self statements. We then hold the quoted text fixed and change only its label from self to peer. This matched-content control yields a $-0.17$pp broken-rate effect (95\% CI $[-0.68,0.35]$), directly showing that the self/peer label does not drive the effect.

This label insensitivity motivates a low-cost attack: an out-of-panel model generates a confident, plausible, but factually \emph{wrong} statement targeting the same probe question and presents it as ``another panel member's independent judgment'' (C7). These generated statements overturn originally-correct verdicts 1.5--2.7$\times$ more often than the naturally occurring wrong peer statements sampled in C6, with a pooled bootstrap 95\% CI excluding parity. This is a comparison of the two tested statement populations, not a content-matched provenance intervention: their selection, length, specificity, and style may contribute to the ratio. A lexicon analysis (Technical Appendix~G) likewise does not isolate a single stylistic mechanism after length normalization. A follow-up black-box experiment iteratively refines C7 statements against a held-out proxy judge and raises damage by $1.28\times$ (95\% CI $[1.16, 1.41]$), showing that the fixed prompt is less damaging than the tested adaptive refinement.

We do not stop at exposing the vulnerability. We design and validate a lightweight defense -- panel-consensus verification: before applying any quoted ``reference opinion'' to a judge's final decision, we first check it for consistency against the majority position implied by the other panel members' genuinely, independently collected judgments on that same sample; if the two disagree, the quoted statement is discarded and the judge falls back on its unanchored (Condition-1) verdict. At scale, this mechanism blocks 84.9\% of fabricated attacks while misclassifying only 4.9\% of genuine peer statements (leave-one-out re-verified; Technical Appendix~B), cutting fabricated attacks' net harm from $-20.3$ to $-0.50$pp (97.5\% reduction), while leaving genuine peer information's positive point-estimate value intact ($+0.42$pp $\to$ $+1.91$pp; individual estimates cross zero at this sample size).

Our contribution is threefold, in escalating order:
\begin{itemize}
\item \textbf{Diagnosis}: a panel-level operational taxonomy of multimodal correlated error, separating language-prior-consistent, visual-hallucination-consistent, and visual-anchoring-failure-consistent patterns.
\item \textbf{Exposing a vulnerability}: controlled contrasts showing that judges are comparably anchored under self and peer framing and readily accept fabricated peer information; under the tested construction, generated wrong quotes are more damaging than sampled genuine errors.
\item \textbf{A validated remedy}: panel-consensus verification, a lightweight defense that checks a quoted judgment against the panel's independently collected blind votes.
\end{itemize}

\section{Related Work}

\paragraph{Correlated errors in LLM judge panels.} LLM-as-Judge is now widely used for automated evaluation. Recent work has begun to examine correlated error within judge panels: \citet{kohli2026nine} find that a panel of nine judges carries the effective, independent information of roughly two, and that supplying the panel with correct answers as a verifiable signal closes at most 11\% of the resulting gap. These studies are confined to text-only settings. We extend the correlated-error question to multimodal judge panels and provide a panel-level operational taxonomy of their error patterns.

\paragraph{Failure modes of VLM-as-Judge.} The past year has seen a wave of studies on the limitations of individual multimodal judges: \citet{li2024vlrewardbench} systematically categorize VL-GenRM error types and find that basic visual perception, rather than reasoning, is the dominant bottleneck; \citet{lee2026mmjudgebias} define nine compositional bias types and systematically probe modality neglect and misalignment; \citet{park2026perceptualjudgment} distinguish ``insufficient perception'' from ``response anchoring'' as separate failure modes and motivate our Type-B/C operational taxonomy; \citet{kumar2026mj1} argue that multimodal judges struggle to ground decisions in visual evidence; and \citet{zou2026informativeness} show that judges can favor more informative-sounding answers over correct ones. To our knowledge, this literature focuses on individual judges rather than correlated error and untrusted peer information in a collaborative-review panel -- the setting studied here.

\paragraph{Adversarial robustness of judges and reward models.} A neighboring line of work studies adversarial manipulation of the \emph{image itself} presented to LVLM judges -- e.g., \citet{hwang2025fooling} show pixel-level perturbations that inflate scores, and \citet{wang2026adversarial} propose a systematic adversarial-robustness evaluation framework for multimodal LLM judges. We study a different attack surface: the quoted, text-level ``reference judgment.'' Conditional on access to that orchestration channel (Section 3.5), the attack requires no image manipulation, only a short natural-language quote.

\section{Method}

\begin{figure*}[t]
\centering
\includegraphics[width=0.90\textwidth]{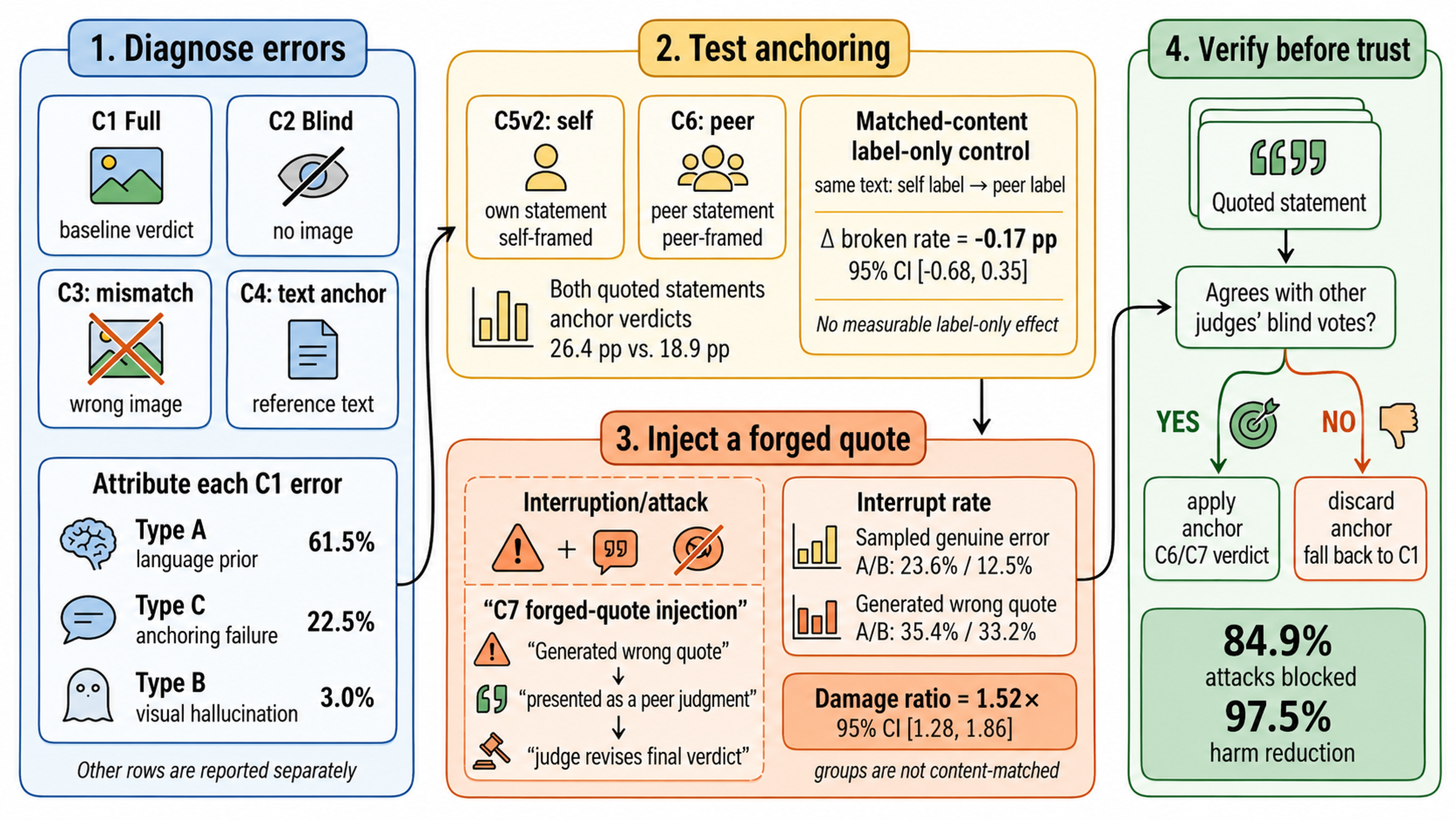}
\caption{The paper's diagnosis-to-defense story. \textbf{Diagnosis} (blue) assigns operational labels to C1 errors using C2--C4. \textbf{Anchoring} (amber) compares self-framed C5v2 with peer-framed C6, then holds content fixed in a label-only control. \textbf{Fabrication} (red) compares generated wrong quotes with sampled genuine peer errors; these groups are not content-matched. \textbf{Defense} (green) accepts a quote only when it agrees with the other judges' blind votes.}
\label{fig:overview}
\end{figure*}

Figure~\ref{fig:overview} summarizes the full design: diagnosis (C1--C4), anchoring tests (C5v2/C6 plus the matched-content label control), fabricated-statement injection (C7), and panel-consensus verification. Verbatim prompt templates, decoding settings, and the defense's gating rule are given in Technical Appendix~E.

\paragraph{3.1 Problem setup and judge panel.} Given an image $I$, a question $q$, and two candidate responses $\{r_A, r_B\}$ (one of which is human-annotated as preferred), the task is for a panel of VLM judges $\mathcal{J} = \{J_1, \ldots, J_7\}$ to decide which response is better. The panel spans seven vendors: gpt-4.1, claude-haiku-4-5, GLM-5V-Turbo, doubao-seed-1.8, grok-4.1-fast-reasoning, Qwen3.5-122B-A10B, and gemini-2.5-flash-lite. The supplementary artifact contains sampled item IDs, generated statements, raw per-judge outputs, cost/failure logs, and analysis/bootstrap scripts.

\paragraph{3.2 Operational error taxonomy (Type A/B/C).} For each incorrect C1 verdict, we first ask whether the same judge is also wrong without the image (C2); if so, it is Type A (language-prior-consistent), irrespective of the probe. Only when C2 is correct do we consult the judge's blind answer to a verified probe at the candidates' point of disagreement: a wrong answer is Type B (visual-hallucination-consistent), while a correct answer followed by a wrong final verdict is Type C (visual-anchoring-failure-consistent). Rows without independently cross-verified probe ground truth are excluded from B/C. This precedence makes the groups mutually exclusive, but the labels remain operational correlates, not identification of a unique causal source (Technical Appendix~C).

\paragraph{3.3 Baseline conditions (C1--C4).} C1 (full) is the standard setting: image + question + both candidate responses. C2 (blind) omits the image. C3 (mismatched image) supplies a randomly chosen, unrelated image, serving as a sanity check for whether the panel genuinely relies on image content. C4 (text-anchored) additionally supplies a reference-answer sentence generated by a strong out-of-panel model, replicating the Nine-Judges-style verifiable-signal repair design, and serves as this paper's most important baseline.

\paragraph{3.4 Anchoring tests: C5v2, C6, and matched content.} C5v2 and C6 share a blind-first structure: (i) without candidate responses, a judge independently answers a probe targeting the candidates' actual point of disagreement; (ii) a blind statement is quoted before the final verdict. C5v2 quotes the judge's \emph{own} statement under self framing. C6 quotes \emph{another} panel member's statement under peer framing (assigned by a deterministic hash of sample ID). Thus C5v2--C6 tests whether a genuine peer statement can anchor, but it varies both content and label. To isolate the label, our matched-content control quotes the exact C5v2 text to the same judge under the C6 peer label; Section~4.4 and Technical Appendix~H report the paired result.

\paragraph{3.5 C7 fabricated-statement injection and threat model.} C7 uses the same peer framing as C6, but substitutes a statement generated by an out-of-panel model: C7-wrong is deliberately plausible and factually wrong; C7-right is generated but correct. C7 measures panel susceptibility to these generated quotes; because C6 and C7 contents are not matched, their damage comparison does not isolate authenticity alone. The attacker is assumed able to alter the quoted-reference field delivered to a judge, as could occur through a compromised orchestration layer, API wrapper, cache, or untrusted upstream agent. We do not claim that ordinary candidate text alone grants this access. Main C7 uses one fixed prompt; a black-box refinement against a held-out proxy raises damage by 1.28$\times$ (95\% CI $[1.16,1.41]$; Technical Appendix~A).

\paragraph{3.6 Defense: panel-consensus verification.} Before applying a C6/C7-style anchoring procedure, we first check the quoted statement for consistency against the majority position implied by the genuinely, independently collected blind statements of the other judges on that panel for that sample; if inconsistent, the anchoring is discarded and the judge falls back on its own unanchored (C1-style) verdict. This uses signals the panel itself produced to verify an externally quoted reference, requiring no annotation or oracle. It significantly outperforms two naive baselines (no defense; a confidence-language heuristic) on catch rate and harm reduction, and beats a maximally conservative ``trust nothing'' baseline on net utility at the point-estimate level (Technical Appendix~B).

\section{Experiments}

\paragraph{4.1 Datasets and sampling.} Dataset A uses VL-RewardBench \citep{li2024vlrewardbench}, a human-annotated vision-language reward-model evaluation benchmark; we stratify-sample and oversample the ``hallucination'' category to ensure sufficient Type-B coverage, obtaining full 7-judge coverage on 1{,}100 samples (1{,}142 C1 baseline errors). Dataset B uses RLAIF-V-Dataset \citep{yu2024rlaifv} (whose $\{\text{image}, \text{question}, \text{chosen}, \text{rejected}\}$ structure naturally corresponds to Dataset A), obtaining full coverage on 179 samples (314 baseline errors). Pooling both datasets yields 1{,}456 baseline errors, used for the panel-accuracy and gap-repair analysis (Section 4.3). The Type A/B/C attribution (Section 4.2) uses a different, larger population: 1{,}075 attribution-eligible Dataset-A rows (excluding those lacking a valid C2 comparison), plus 557 Dataset-B rows after supplementing with 120 hallucination-focused samples for Type-B coverage (Technical Appendix~C); pooling gives 1{,}632 rows for the Type A/B/C decomposition. (During dataset selection we evaluated MM-JudgeBias \citep{lee2026mmjudgebias} and MJ-Bench \citep{chen2024mjbench} as candidate second datasets and rejected both due to task-format incompatibility; see Technical Appendix~D.)

\paragraph{4.2 Error-taxonomy results.} Pooling both datasets after the verification gate, Type A accounts for 61.5\% (1{,}004), Type C for 22.5\% (368), and Type B for 3.0\% (49); 8.6\% (140) lack cross-verified probe ground truth and 4.4\% (71) have no probe question (all five groups sum to 1{,}632). Under this operational taxonomy, the Type-C pattern is $7.5\times$ as frequent as Type B. The ratio is descriptive rather than causal and inherits the small Type-B cell and validator-dependent pseudo-ground truth documented in Limitations and Technical Appendix~C. Per-dataset A/C/B shares are 59.7\%/22.4\%/3.8\% and 65.0\%/22.8\%/1.4\%.

\paragraph{4.3 Panel accuracy and gap-repair comparison.}

\begin{table}[t]
\centering
\small
\setlength{\tabcolsep}{4pt}
\begin{tabular}{@{}lccc@{}}
\toprule
Condition & Dataset A & Dataset B & Pooled repair \\
\midrule
C1 (full/baseline) & 80.0\% & 58.1\% & --- \\
C2 (blind) & 67.9\% & 52.1\% & --- \\
C3 (mismatched image) & 77.3\% & 49.4\% & --- \\
C4 (text-anchored) & 82.0\% & 61.3\% & 35.1\% \\
C5v2 (own statement) & 79.5\% & 65.6\% & 36.9\% \\
C6 (peer statement) & 81.5\% & 66.3\% & \textbf{44.6\%} \\
\bottomrule
\end{tabular}
\caption{Panel accuracy by condition and pooled gap-repair rate on C1 baseline errors.}
\label{tab:accuracy}
\end{table}

The sanity check C1 $>$ C3 $>$ C2 holds pooled (77.6\%/74.9\%/66.0\%) and on Dataset A, confirming the panel relies on real image content there; it does not hold on Dataset B, where C3 (49.4\%) is slightly below C2 (52.1\%), consistent with Dataset B's smaller, noisier sample. C6's gap-repair rate is consistently higher than both C4 and C5v2 on both datasets (45.2\% / 42.4\%), one of this study's most robust quantitative findings.

\paragraph{4.4 Net effect and anchoring mechanism.} We report sample-level cluster-bootstrap 95\% confidence intervals (8{,}000 replicates, resampling all judge rows for a sample together). Pooled, C4's net effect is $+1.89$pp (95\% CI $[0.70,3.07]$), C6's is $+1.05$pp ($[-0.66,2.78]$), and C5v2's is $-0.15$pp ($[-1.48,1.25]$). Only C4 is statistically distinguishable from zero. C6's point estimate is positive on each dataset, whereas C5v2 is slightly negative on Dataset A and positive on Dataset B (Table~\ref{tab:accuracy}). C6's advantage is clearer in gap repair and the broken/kept diagnostic below, both computed on the larger population of C1-baseline-error rows.

\begin{table}[t]
\centering
\small
\begin{tabular}{@{}lcc@{}}
\toprule
Condition & Net eff., pp [95\% CI] & Gap, pp [95\% CI] \\
\midrule
C4 & $+1.89$ $[0.70,3.07]$ & --- \\
C5v2 (own) & $-0.15^{\dagger}$ $[-1.48,1.25]$ & 26.4 $[20.5,32.5]$ \\
C6 (peer) & $+1.05^{\dagger}$ $[-0.66,2.78]$ & 18.9 $[12.4,25.9]$ \\
\bottomrule
\end{tabular}
\caption{Net effect on panel accuracy and broken-vs-kept correctness gap of the quoted statement, pooled across both datasets; sample-level bootstrap 95\% CIs in brackets. $\dagger$: CI crosses zero.}
\label{tab:mechanism}
\end{table}

Splitting ``verdict broken'' and ``verdict kept'' by whether the quote was correct (Table~\ref{tab:mechanism}), C5v2 shows a 26.4pp gap (95\% CI $[20.5,32.5]$) and C6 an 18.9pp gap (95\% CI $[12.4,25.9]$). Both exclude zero: a genuine peer statement clearly anchors, not only a judge's own prior answer. Their difference is $+7.5$pp (95\% CI $[1.9,13.0]$), so self anchoring is modestly stronger. Because C5v2 and C6 also quote different content, this comparison alone cannot attribute that difference to source labeling. The matched-content control supplies the clean test: across 5{,}406 matched rows (4{,}212 with C1 originally correct for the broken-rate contrast), relabeling the identical own statement as a peer's changes the broken rate by $-0.17$pp (95\% CI $[-0.68,0.35]$). Together, the results reject a self-label-specific explanation and support general anchoring toward quoted visual evidence.

\begin{figure}[t]
\centering
\includegraphics[width=\linewidth]{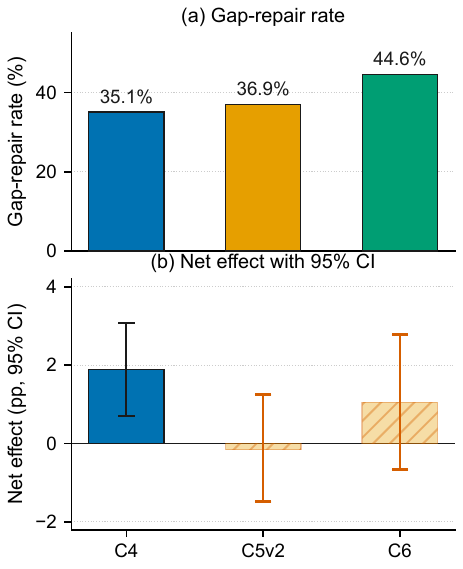}
\caption{Gap-repair rate (top) and net effect on panel accuracy (bottom, with sample-level bootstrap 95\% CI error bars) for C4, C5v2, and C6, pooled across both datasets. C6 leads on gap-repair, the metric computed on the larger C1-baseline-error sample. On net effect, only C4's 95\% CI excludes zero; C5v2's and C6's both cross zero at the current sample size and are drawn with hatched bars to flag this explicitly (Section 4.4) -- net effect should not be read as a confident point estimate for C5v2 or C6.}
\label{fig:core_results}
\end{figure}

\paragraph{4.5 C7 vulnerability validation.} Generated wrong quotes overturn 35.4\% of originally-correct verdicts on Dataset A ($n=4168$) and 33.2\% on Dataset B ($n=500$), versus 23.6\% ($n=502$) and 12.5\% ($n=38$) for sampled genuine wrong peer statements. The pooled differential-damage ratio is 1.52$\times$ (sample-level bootstrap 95\% CI $[1.28,1.86]$); per-dataset ratios are 1.50$\times$ $[1.25,1.84]$ and 2.66$\times$ $[1.05,6.86]$. These intervals establish a difference between the tested populations, not that fabrication provenance causes it: C7 is generated to be concise, plausible, and confident, whereas C6 errors occur naturally. Susceptibility spans 17.1--47.9\% across all seven judges, so the pooled result is not driven by one model (Technical Appendix~F). Dataset B remains imprecise, and the C7-right repair direction is not cross-dataset-stable.

\begin{table}[t]
\centering
\small
\begin{tabular}{@{}lcc@{}}
\toprule
Quoted statement & Dataset A & Dataset B \\
\midrule
Genuine wrong peer & 23.6\% ($n{=}502$) & 12.5\% ($n{=}38$) \\
\textbf{Fabricated (C7-wrong)} & \textbf{35.4\%} ($n{=}4168$) & \textbf{33.2\%} ($n{=}500$) \\
\midrule
Damage ratio [95\% CI] & 1.50 $[1.25,1.84]$ & 2.66 $[1.05,6.86]$ \\
\bottomrule
\end{tabular}
\caption{Rate at which a quoted statement overturns an originally-correct verdict for sampled genuine errors versus generated wrong quotes; the groups are not content-matched. Sample-level bootstrap 95\% CIs are in brackets; pooled differential-damage ratio: 1.52 (95\% CI $[1.28,1.86]$).}
\label{tab:vulnerability}
\end{table}

\paragraph{4.6 Defense mechanism effectiveness.} On Dataset A (large sample, most reliable), panel-consensus verification blocks 84.9\% of C7 fabricated attacks while misclassifying only 4.9\% of genuine correct peer statements and 15.1\% of fabricated correct statements. These are leave-one-out (LOO) re-verified figures, excluding each gated judge's own vote from the majority signal it is checked against to rule out a self-inclusion concern (Technical Appendix~B). Pre-verification bootstrap estimates were nearly identical (87.0\% $[84.6\%,89.3\%]$; 3.8\%; 13.0\%), so we report LOO as the headline and the bootstrap-CI'd figures as a robustness check (Table~\ref{tab:defense}). With the defense applied, C7-wrong's net harm falls from $-20.3$pp to $-0.50$pp, a 97.5\% reduction (pre-verification: $-0.19$pp/99.1\%, 95\% CI $[94.2\%, 99.9\%]$). C6's positive point-estimate net effect does not shrink under the defense ($+0.42$pp $\to$ $+1.91$pp LOO; $+1.34$pp pre-verification). The pre-verification C6 estimates individually cross zero under their own 95\% CI at the current sample size, though, so ``improves'' should be read as directional, not a statistically confirmed gain -- the catch-rate and harm-reduction numbers are the defense's more robust claims. Each sample's gating decision depends only on that sample's own panel votes, independently of every other sample, so this is a per-sample-independent simulated deployment check, not production-grade real-time deployment testing. Dataset B (small sample) shows the same direction but with noisier magnitudes.

\begin{table}[t]
\centering
\small
\setlength{\tabcolsep}{3pt}
\begin{tabular}{@{}lcc@{}}
\toprule
Condition (pp), LOO & No defense & With defense \\
\midrule
C6 net & $+0.42$ & $+1.91$ \\
C7-wrong net & $-20.3$ & $\mathbf{-0.50}$ \\
\bottomrule
\end{tabular}
\caption{Net effect (pp) before/after panel-consensus verification (Dataset A, $n{\approx}2831$--$4391$), leave-one-out (LOO) re-verified point estimates (Technical Appendix~B). Pre-verification bootstrap estimates (95\% CI) agree in direction: C6 $+0.42^{\dagger}[-1.26,2.18]\to+1.34^{\dagger}[-0.35,3.04]$pp; C7-wrong $-20.3[-22.9,-17.6]\to-0.19[-1.16,0.80]$pp; catch rate 87.0\% $[84.6\%,89.3\%]$ (vs.\ LOO 84.9\%). $\dagger$: pre-verification CI crosses zero.}
\label{tab:defense}
\end{table}

\begin{figure*}[t]
\centering
\includegraphics[width=0.92\textwidth]{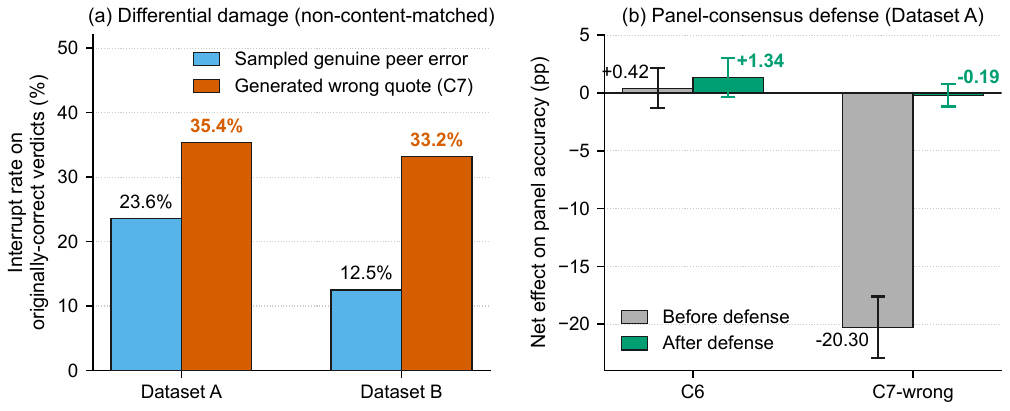}
\caption{(a) Under the tested, non-content-matched constructions, generated C7-wrong quotes interrupt originally-correct verdicts more often than sampled genuine wrong peer statements (pooled ratio 1.52$\times$, 95\% CI $[1.28,1.86]$). (b) On Dataset A, panel-consensus verification reduces C7-wrong's net harm by 99\% ($-20.3$pp $\to$ $-0.19$pp; pre-verification bootstrap estimates shown with 95\% CI bars); leave-one-out re-verification gives a consistent 97.5\% reduction ($-20.3$pp $\to$ $-0.50$pp). The C6 utility estimates cross zero, so the panel shows no detected benign-signal trade-off rather than a confirmed improvement.}
\label{fig:vuln_defense}
\end{figure*}

\section{Discussion}

Our findings compose a diagnosis-vulnerability-remedy arc. \textbf{Diagnosis}: an operational taxonomy separates error patterns that respond differently to intervention. \textbf{Vulnerability}: genuine peer statements anchor, changing only the self/peer label has no measurable pooled effect, and generated wrong quotes cause greater damage than the sampled genuine-error population. The last comparison is differential, not provenance-only causal. \textbf{Remedy}: the panel's independent blind judgments can verify a quote without an external oracle. Panel-consensus verification nearly eliminates C7's harm; its C6 utility result remains a favorable but statistically inconclusive point estimate.

\section{Limitations}

(1) Type B is small (49 verified rows, 41 from Dataset A) despite a targeted Dataset-B expansion, and its pseudo-ground-truth reproducibility is validator-dependent (third-model agreement 46.0\%, human agreement 69.8\%; Technical Appendix~C). The Type A/B/C labels are therefore operational patterns, not identified causal sources. (2) C5v2's and C6's net effects on panel accuracy are not statistically distinguishable from zero. (3) Dataset B has only 179 samples; several secondary effects are noisy, including its damage ratio ($[1.05,6.86]$) and C7-right repair direction. (4) C4/C7 reference answers and the Type-B/C split depend on model-generated pseudo-ground truth, although we cross-verify probes, exclude unverified rows, and report independent model and human audits. (5) The defense is evaluated in a per-sample simulated deployment and assumes a trustworthy majority; coordinated quote injection or a panel-wide blind spot may defeat it, and panel-size sensitivity remains untested. (6) C7 generated statements and genuine C6 errors are not content-, length-, style-, or error-type-matched. Their 1.52$\times$ ratio is a differential-damage estimate for these populations, not a provenance-only effect. The two-round proxy-judge refinement increases damage by 1.28$\times$ over fixed C7, but does not remove this confound or represent the strongest adversary. (7) We use sample-level cluster-bootstrap 95\% CIs (8{,}000 replicates) for core comparisons, resampling all judge rows for a sample together. We do not report formal CIs for taxonomy proportions, per-judge breakdowns, C7-right repair direction, or leave-one-out point estimates.

\section{Conclusion}

This paper studies the reliability and security of a collaborative-review design in multimodal LLM judge panels: consulting another member's independent visual judgment. We find that: (1) an operational taxonomy separates language-prior-, visual-anchoring-failure-, and visual-hallucination-consistent patterns; (2) peer statements anchor strongly, while matched-content relabeling shows no measurable pooled self/peer label effect; (3) generated wrong quotes are a low-cost attack and, under the tested non-matched construction, damage more verdicts than sampled genuine errors; and (4) panel-consensus verification effectively mitigates the attack without a detected loss of benign collaborative signal. Future work should test content-matched authenticity interventions, coordinated attacks, panel-size sensitivity, and live deployment.

\bibliographystyle{plainnat}
\bibliography{references}

\end{document}